\documentclass{amia}
\pdfoutput=1
\usepackage{graphicx}
\usepackage[labelfont=bf]{caption}
\usepackage[superscript,nomove]{cite}
\usepackage{color}

\usepackage{placeins}

\usepackage{subcaption}
\usepackage{graphicx}

\begin{document}

\title{Quantification of BERT Diagnosis Generalizability Across Medical Specialties Using Semantic Dataset Distance
}

\author{Mihir P. Khambete$^{1,2}$,  William Su, MD$^{1,3}$, Juan C. Garcia, PhD$^{1}$, Marcus A. Badgeley, PhD$^{1,4}$}

\institutes{
    $^1$ nference LLC, Cambridge, MA;
    $^2$ Department of Electrical Engineering and Computer Science, Massachusetts Institute of Technology, Cambridge, MA;
    $^3$ Department of Radiation Oncology, Penn Medicine, University of Pennsylvania Health System, Philadelphia, PA;
    $^4$ Department of Brain and Cognitive Sciences, Massachusetts Institute of Technology, Cambridge, MA\\
}

\maketitle

\noindent{\bf Abstract}

\textit{Deep learning models in healthcare may fail to generalize on data from unseen corpora. Additionally, no quantitative metric exists to tell how existing models will perform on new data. Previous studies demonstrated that NLP models of medical notes generalize variably between institutions, but ignored other levels of healthcare organization. We measured SciBERT diagnosis sentiment classifier generalizability between medical specialties using EHR sentences from MIMIC-III. Models trained on one specialty performed better on internal test sets than mixed or external test sets (mean AUCs 0.92, 0.87, and 0.83, respectively; p = 0.016).  When models are trained on more specialties, they have better test performances (p $<$ 1e-4).  Model performance on new corpora is directly correlated to the similarity between train and test sentence content (p $<$ 1e-4). Future studies should assess additional axes of generalization to ensure deep learning models fulfil their intended purpose across institutions, specialties, and practices.
}

\section*{Introduction}
Natural Language Processing (NLP) models could enhance clinical research and patient care by automatically curating electronic health record (EHR) notes.  The state-of-the-art NLP models are high capacity deep neural networks, and other deep neural networks (for image recognition) have been shown to be particularly prone to overfitting in healthcare contexts (Badgeley et al. 2019; Zech et al. 2018)\cite{ref11,ref12}.  Generalizability is the capacity of a model to perform comparably on test sets derived from domains not used for model training.  NLP models that better generalize would enable models trained on existing labelled datasets to be deployed to new research questions and EHRs.

A major reason for the generalizability issue is that medical notes are largely entered as free text, and therefore lack the organization typical of structured data in the EHR. Different hospital systems may use different note styles, as can different medical specialties or caregivers within an institution. Thus, any generalizable NLP model must make accurate predictions on a wide range of note styles after training on medical notes from a small number of institutions. 

Current approaches to work around the generalizability problem have included using domain adaptation and model retraining, which must be applied each time a model is used on data from a new hospital system (Wu et al, 2014)\cite{ref10}. This precludes large-scale application of NLP for medical note curation. 

Some prior studies have tested how well NLP models generalize across healthcare institutions.  
Cancer diagnosis has been shown to be generalizable when using training data from multiple hospitals, at the expense of a slight loss in performance on test sets derived from hospitals used for model training (Santus et al. 2019)\cite{ref7}. Similarly, temporal reasoning can also be considered generalizable when a diverse training set is used (Velupillai et al. 2015)\cite{ref8}. Another group trained a model to predict the prognosis of ICU patients at an academic medical center, and found that results generalized better to another academic medical center than a local community hospital (Marafino et al. 2018)\cite{ref5}. On the other hand, existing models for negation detection have not generalized well across institutions. Negex, a simple regular expression based algorithm, fails to detect terms such as ``afebrile" which indicate negation but are used less frequently in some hospital systems (Wu et al. 2014)\cite{ref10}. However, these studies have only assessed generalizability across institutions.

It remains unclear whether NLP models are generalizable on finer axes of generalization such as medical specialty and the type of EHR note.  In this study, we used MIMIC-III, an ICU EHR dataset from Beth Israel Deaconess Medical Center (Johnson et al. 2016)\cite{ref3}. We trained state-of-the-art NLP models (Bidirectional Encoders, BERT)\cite{ref2} to classify the sentiment of possible diagnoses across three medical specialties. Using training and test sets containing sentences from either one, two, or all three medical specialties, we answer questions about NLP generalizability at the medical specialty level. We investigated the effect of increasing the training set diversity, defined as the number of specialties represented in our study. Additionally, we examined how semantic similarity between a training set and a test set is correlated to model performance.

\section*{Methods}
\subsection*{Dataset}
Our study used both structured data and text from MIMIC-III, an EHR dataset from the Beth Israel Deaconess Medical Center (BIDMC) critical care units between 2001 and 2012. We primarily used the NOTEEVENTS table which contains over 2 million medical notes’ text and metadata fields such as note type (e.g ECG, Nursing, Discharge Summary). 

Our main objective was to assess whether an NLP model could generalize to sentences from medical specialties other than the one(s) which the model was trained on.  We performed 2 types of analyses 1) unsupervised clustering of the EHR  note embeddings and 2) supervised modeling of diagnosis sentiment for three specialties.  We randomly sampled 5,020 ``background" notes and fetched 2,955 sentences containing a medical diagnosis term.  The unsupervised analysis used both background sampling and sentences containing diagnoses, whereas model training and testing only involved sentences containing diagnoses.  

We identified sentences containing disease names from three medical specialties: oncology, cardiology, and pulmonology.  These specialties were selected because of the high prevalence of disease in critical care patient populations. Disease tokens for these specialties were identified from the NIH TCGA project and the International Classification of Diseases-10 (ICD-10) (https://www.cancer.gov/tcga and World Health Organization, 2019 respectively).  Disease names for cardiology and pulmonology were simplified to match shorthand commonly used by clinicians in EHR notes.  For example, in our cardiology disease names, we selected the simplified term ``heart failure" instead of specific variants such as ``systolic (congestive) heart failure", ``right heart failure", and ``end stage heart failure".  The most common disease names found for each specialty are included in Supplementary Table 2.

\subsection*{Language Processing}
Each medical note was broken into sentences using the SpaCy English sentencizer. Sentences were then broken into words (excluding punctuation characters). Continuous stretches of 1-6 words were matched with the disease names from a medical specialty, to account for disease names containing several words e.g. cardiac arrest or chronic obstructive pulmonary disease. If a match was found, the sentence, name of the disease, and the associated note metadata were collected. Sentences with more than 512 tokens were excluded, since our model architecture had a limit of 512 tokens per sentence. Additionally, we ensured that all sentences were unique.

\subsection*{Unsupervised Techniques}
We used a deep bidirectional encoder model (BERT) to create language embeddings (Devlin et al. 2018)\cite{ref2} The variant we used is called SciBERT, which was pre-trained on text from scientific publications (Beltagy, Lo, and Cohan 2019)\cite{ref1} and is available in the python transformers package (Wolf et al. 2019)\cite{ref9}.  The transformer generates embeddings for individual words. To compute sentence or document embeddings we apply mean pooling as previously described and implemented in the sentence-transformers package (Reimers and Gurevych 2019)\cite{ref6}.

BERT embeddings of sentences and documents are 768-dimensional numerical representations of the texts’ semantic content. We projected these embeddings into 2-dimensional spaces using t-distributed Stochastic Neighbor Embedding (tSNE, (Maaten and Hinton 2008)\cite{ref4} and Principal Component Analysis (PCA) for unsupervised analyses. tSNE and PCA projections were generated and visualized using the scikit-learn and plotly packages. tSNE was used to identify local clusterings of notes; we used a perplexity value of 15 when creating tSNE projections to allow separation of points into separate clusters when visualized. PCA was used to retain distances between points and project queried sentences into an unbiased background distribution.  The PCA rotation was fit on 1,004 randomly sampled sentences from MIMIC-III, and then the learned rotation was applied to the 2,955 specialty sentences.

Semantic similarity between sentences was measured by computing the cosine distance between pairs of sentences’ 768-dimensional embeddings.  The similarity of 2 data partitions was estimated by measuring the cosine distance between all pairs of sentences and computing the median: the Median Cosine Distance (MCD).  Median rather than mean cosine distance was used to mitigate the effect of outliers.

\subsection*{Supervised Model Development}
\subsubsection*{Sentence Annotation}
Sentences with diagnoses were annotated by one of two authors: a resident physician (WS) or a physician scientist (MAB). Given a sentence and the disease contained in that sentence, the annotator marked the sentence as one of three ground truth labels: ``Yes" (indicating the patient had the disease, or if a fetus had the disease in obstetric cases), ``No" (if the patient did not have the disease), or ``Maybe" (if there was insufficient evidence to exclude or confirm the disease). 

An example ``Yes" sentence is ``pt was found to have \textbf{cardiomyopathy} and is in heart failure" , since the patient has a positive diagnosis for cardiomyopathy. An example ``No" sentence is ``No episodes of \textbf{tachycardia} at this time". An example ``Maybe" sentence is ``Some left \textbf{heart failure} cannot be excluded since the patient is supine"; there is insufficient evidence to mark this sentence as either positive or negative for heart failure.  If a sentence contained information for a family member (which is routinely collected as part of the medical history), the sentence was marked as ``No" even if the relative had the indication.  

The initial round of labeling saw an overwhelming majority of sentences annotated as ``Yes". To mitigate the class imbalance, we collected additional sentences from MIMIC-III that the preliminary BERT model predicted as ``Maybe" or ``No", and our annotators (WS, MAB) confirmed or corrected each of the model’s predictions.

\subsubsection*{Train-Test Split}
We created seven different combinations of sentences from the 3 specialties - and split each into a train and test set. For each medical specialty, approximately 70\% of sentences were used for training (selected at random) while the rest were used for testing. Each train-test set contained sentences from one, two, or three medical specialties (see Table 1). For datasets containing sentences from multiple specialties, we used roughly the same number of sentences from each of the component specialties while keeping the total number of sentences consistent from train set to train set and from test set to test set. We trained sciBERT classification models on each of the 7 training datasets and then tested each model on all 7 test datasets, yielding 49 different train-test performance tests.
\\

\begin{table}
\centering
\caption{Train and Test Set Compositions by Specialty}
    \begin{tabular}{|l|l|l|l|l|l|l|}
    
    \hline
    \textbf{Medical Specialty Sets} & \multicolumn{2}{|l|}{Oncology}
    & \multicolumn{2}{|l|}{Cardiology}
    & \multicolumn{2}{|l|}{Pulmonology}\\
   \hline
    & Train & Test & Train & Test & Train & Test \\ \hline
              Oncology Only &      783 &        337 &          0 &          0 &           0 &          0 \\ \hline
            Cardiology Only &        0 &          0 &        631 &        271 &           0 &          0 \\ \hline
           Pulmonology Only &        0 &          0 &          0 &          0 &         653 &        280 \\ \hline
    Oncology and Cardiology &      392 &        168 &        316 &        136 &           0 &          0 \\ \hline
   Oncology and Pulmonology &      392 &        168 &          0 &          0 &         326 &        140 \\ \hline
 Cardiology and Pulmonology &        0 &          0 &        316 &        136 &         326 &        140 \\ \hline
      All Three Specialties &      261 &        112 &        210 &         90 &         217 &         93 \\ \hline
    \end{tabular}
\end{table}

\newpage
We define the following relations for train-test set pairs based on whether the same or different medical specialties are included: 1) native, 2) partial, and 3) external. ``Native" performance tests include the same set of specialties in training and test sets.  If there is no overlap between the specialties included in the training and test set, we designate the performance test as ``external". All performance tests with any but not complete overlap are designated ``partial".

\subsubsection*{Model Architecture}
Our model included three modules. The first module was a cased sciBERT transformer which generated 768-dimensional embeddings from an input sentence. The embeddings were fed into a linear module with three output units. Lastly, a 3-class softmax unit generated confidences (probabilities) for each of the three output classes, such that the predicted class was that with the highest confidence.

Training used a categorical cross-entropy loss for each sentence using the ground truth labels and the model outputs from the softmax layer. Our model used a learning rate of 3e-5 in order to preserve the pre-training weights while fine-tuning the model. We chose hyperparameters of dropout rate = 0.9, and used the Adam optimizer along with a warmup schedule and LayerNorm. 10\% of the training data was held out and used as a validation set.
\subsubsection*{Evaluating Model Performance}
We generated receiver operator characteristic (ROC) and precision-recall (PR) curves for each train-test pair on each output class using the scikit-learn package.  The primary performance metric was the area under the ROC curve (AUC) on the epoch with the lowest validation loss.  We evaluated statistical significance of generalizability between medical specialties and generalization by training set diversity using the repeated-measures ANOVA test implemented with the python pingouin package. 

Finally, we analyzed the relationship between AUC and MCD between a train-test set to determine whether similarity between a train-test set is correlated to how well models generalize.  Significant associations between model test performance and MCD were evaluated using a two-tailed t-test implemented with the python statsmodels package.

\section*{Results}
\subsection*{Data Characteristics}
We queried medical notes from the MIMIC dataset to obtain 1) a random subset of 5,020 documents for unsupervised note analysis and 2) 2,955 sentences containing medical diagnosis terms from one of 3 medical specialties to train diagnosis models.  The distribution of note types was unbalanced in both cohorts, and the notes containing oncology diagnoses were notably enriched in discharge summaries (Supplementary Table 1). The medical diagnoses that were found most frequently are shown in Table S4.  

Documents and sentences were split into subword tokens.  Sentences with pulmonology, cardiology, and oncology diagnoses had an average of 43, 48, and 65 tokens per sentence (see full distributions in Figure S10).
\newpage
\subsection*{Unsupervised Analysis of Clinical Text Embeddings}
\begin{figure}[hbt!]
    \centering
    \includegraphics[width=\textwidth]{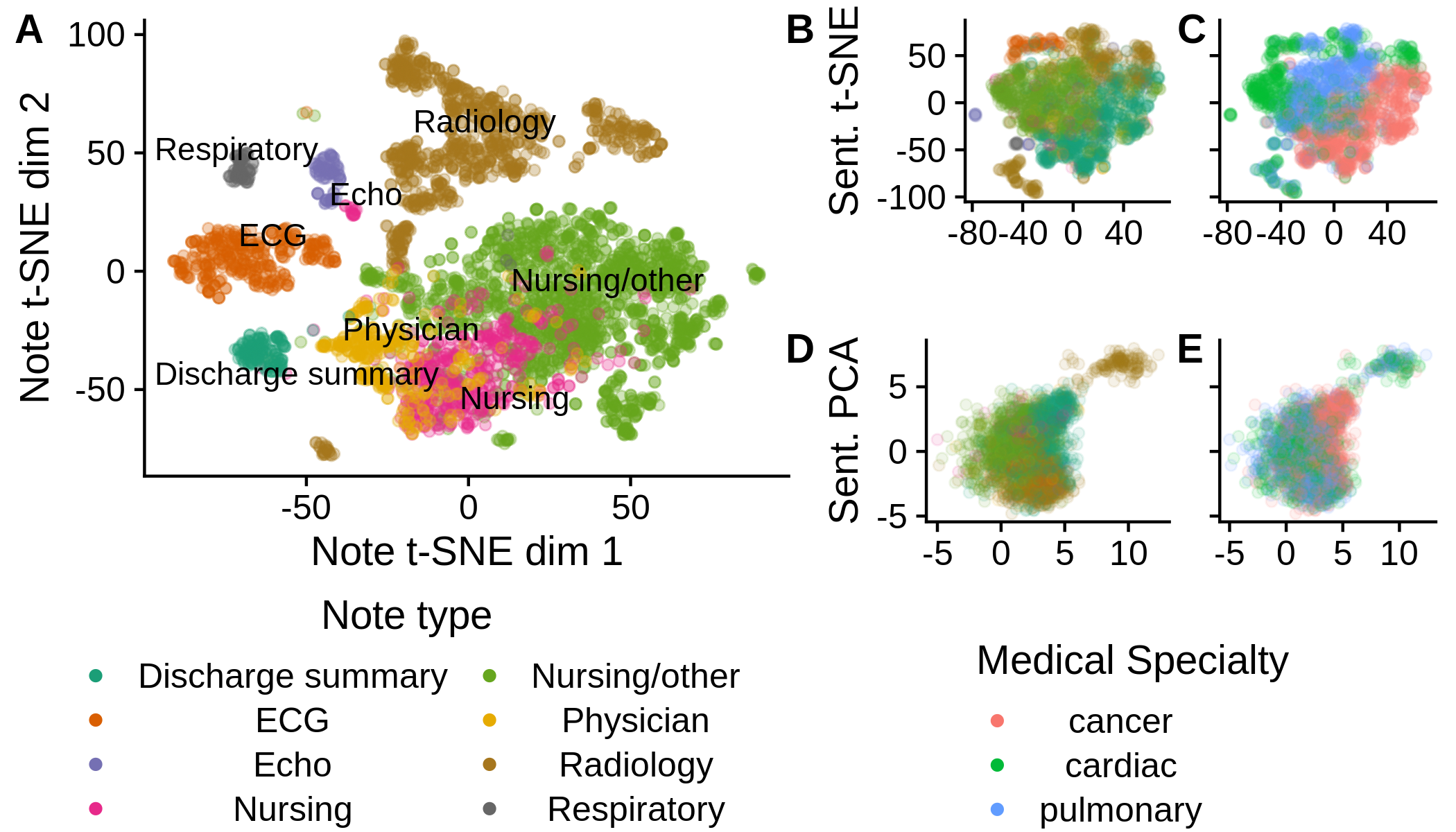}
    \caption{Unsupervised analysis of medical notes.  (A) Document-level tSNE for 5,020 randomly sampled medical notes.  Sentence-level (B, C) tSNE and (D, E) PCA for 2,955 sentences containing diagnoses.  Each point represents a document or sentence that was embedded using a sciBERT model and projected into 2-dimensions with tSNE (A, B, C) and PCA (D, E).  Colors indicate the (A, B, D) type of note (e.g. nursing, physician, pharmacy) and (C, E) diagnosis specialty (cancer, cardiac, pulmonary).}
    \label{fig1}
\end{figure}

Documents were vectorized using the sciBERT model which was pre-trained on scientific corpora.  The tSNE projection of SciBERT embeddings of randomly sampled notes are shown in Figure 1A. Each point, whose xy-coordinates represent the 2-dimensional projection of a document-vector embedding, was subsequently colored according to note type in the document metadata.  Document-vector embeddings cluster into neighborhoods primarily based on the type of note.  General note types were intermixed and overlapping (e.g. nursing and physician notes).  Specialty notes are packed into single isolated clusters (e.g. ultrasound and respiratory notes) or distributed into several subpopulations (e.g. radiology and ECG). For example, the radiology cluster centered roughly around (-50, -75) in the tSNE projection corresponds to notes for patients undergoing catheterization for angiography and other vascular procedures, while the nursing cluster centered around (10, -70) contains notes related to the care of newborns in the ICU.

Sentences containing diagnoses were similarly embedded and visualized with SciBERT.  Like the document tSNE, the sentence clusters coincide with the type of note they came from (Figure 1B); however, the sentence-level clustering is much less pronounced than the document-level clustering.  The specialty coloring (Figure 1C) reveals the disproportionate note types for each specialty. Manual review of sentence clusters reveals content-based neighborhoods. For example, cardiology sentences projected near (-50, 0) discuss arrhythmias such as tachycardia and ventricular fibrillation.

In addition to tSNE, we visualized our sentence embeddings using PCA, which preserves longer distance relations between high-dimensional data points (Figures 1C, D). PCA was fitted using an unbiased distribution of approximately 1,000 sentences sampled from across MIMIC-III.  There is a primary group of sentences with all 3 specialty diagnoses as well as a secondary group of sentences from radiology notes that only contains cardiac and pulmonary sentences.  As in tSNE, we see that sentences are more separated based on note type than the diagnosis token medical specialty.  The bivariate principal component distributions of cardiac and pulmonary specialty sentences appear more similar than the oncology sentence distribution.
\subsection*{Model Performance}
Model performance was assessed by measuring AUC for each train-test pair and label (Supplementary Table 3), as well as by constructing precision-recall curves (Supplementary Figure 2).  

We first investigated whether models could generalize across medical specialties. We summarize test set performance for models trained on a single specialty by macro-averaging AUC across labels and averaging across train-test relationships (native, partial overlap, and external; see methods). We find that AUC monotonically decreases as the overlap between specialties decreases (Figure 2A; repeated measures ANOVA p = 0.0163).
\subsection*{Unsupervised Analysis of Clinical Text Embeddings}
\begin{figure}[hbt!]
    \centering
    \includegraphics[width=\textwidth]{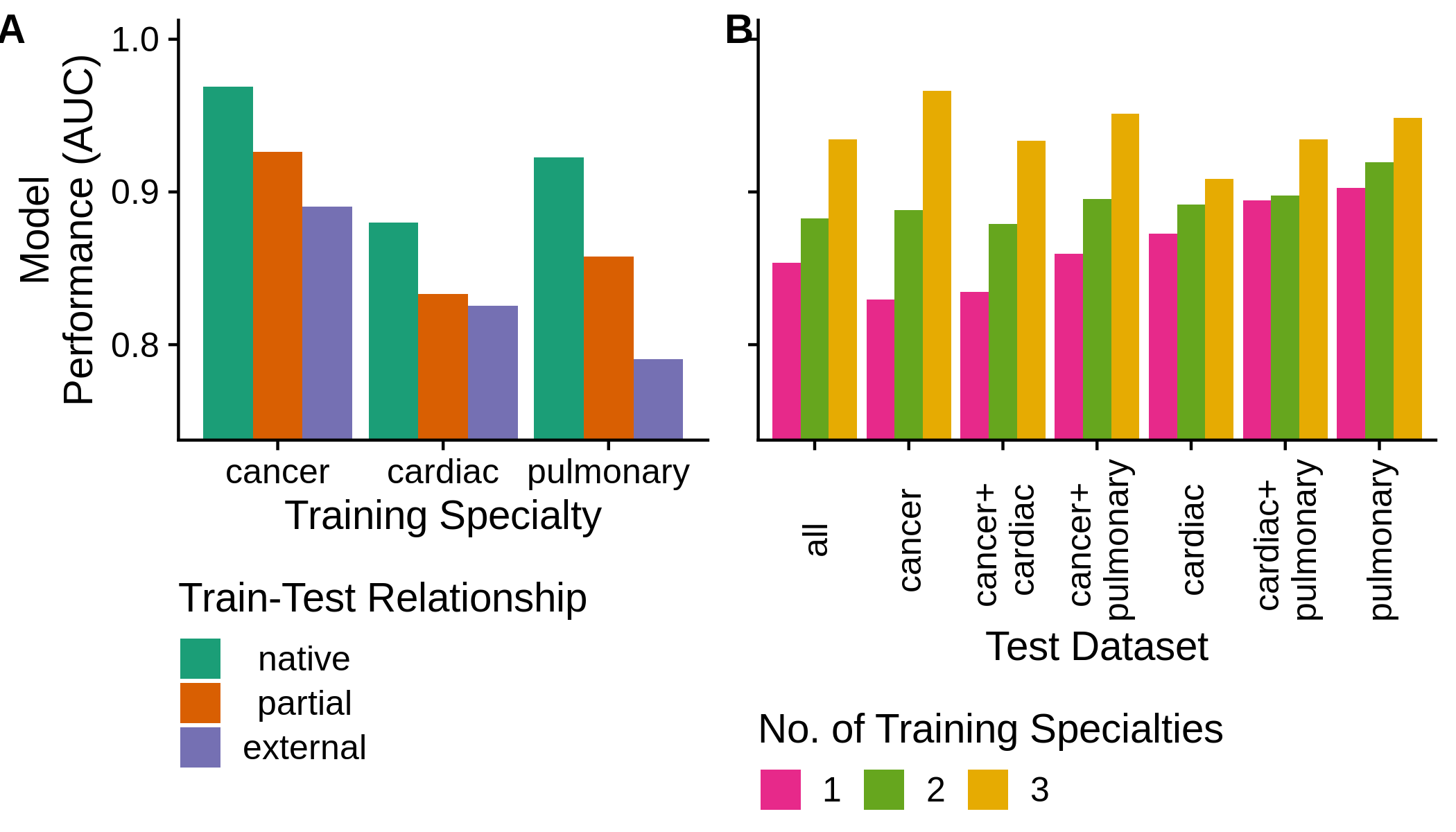}
    \caption{AUC bar plots of model performance tests with different groupings to test primary hypotheses.  (A) AUC vs train-test set relation for each single-disease model, averaged over labels and test sets.  (B) AUC vs how many specialties were used to train the model.  Bars are colored by the number of medical specialty diagnoses used in model training and grouped by test set name, averaged over labels and training specialties.}
    \label{fig2}
\end{figure}

To test whether the diversity of training data improves test performance, we grouped models based on how many specialties they were trained and summarized performance on each test set.  Specifically, we take a macro-average across labels and average the results for models trained on diagnoses from 1, 2, or 3 specialties (Figure 2B).  We found that for every single test set, the average model performance monotonically increased as more training data specialties are used to train the models. Using the repeated-measures ANOVA test, we found that the differences in macro-average AUC were statistically significant (p $<$ 1e-4).

\subsection*{Secondary Performance Evaluation}
We used MCD to measure the semantic similarity between different train-test sets and investigated how this affects model performance.  We measured MCD both between and within every combination of train and test datasets (see supplementary tables 2 and 3). We group dataset pairs using the following relations: intra-dataset, internal, partial, and external (the latter three relations are between training set-test set pairs only).  We found significant differences between the mean MCDs of native, partial, and external train-test pairs (mean MCDs: 0.230, 0.237, and 0.245, respectively; one-way ANOVA p = 5.78e-5).  However, we did not find significant differences between intra-set and native train-test MCDs (mean MCD 0.230 vs 0.231; one-sample t-test p = 0.819).  We validated that MCD is not different for subsamples from the same distribution, and that it progressively increases for train-test sets from 1) same distribution, 2) distribution mixtures, and 3) separate distributions.

Model test AUCs are shown as a function of the MCD for every combination of train-test sets in Figure 3. Using ordinary least-squares (OLS) regression, we find an inverse relationship between MCD and AUC for unequivocal diagnoses (p=6.5e-6 and 4.5e-5 for sentences labelled ``Yes" or ``No" respectively) but no difference when a diagnosis is uncertain (p = 0.70 for ``Maybe") (Table 2).  The median cosine distances and AUCs for each train-test set are provided in Supplementary Table 3. Baseline median cosine distances between each dataset and itself are given in Supplementary Table 2.

\begin{figure}[hbt!]
    \centering
    \includegraphics[width=\textwidth]{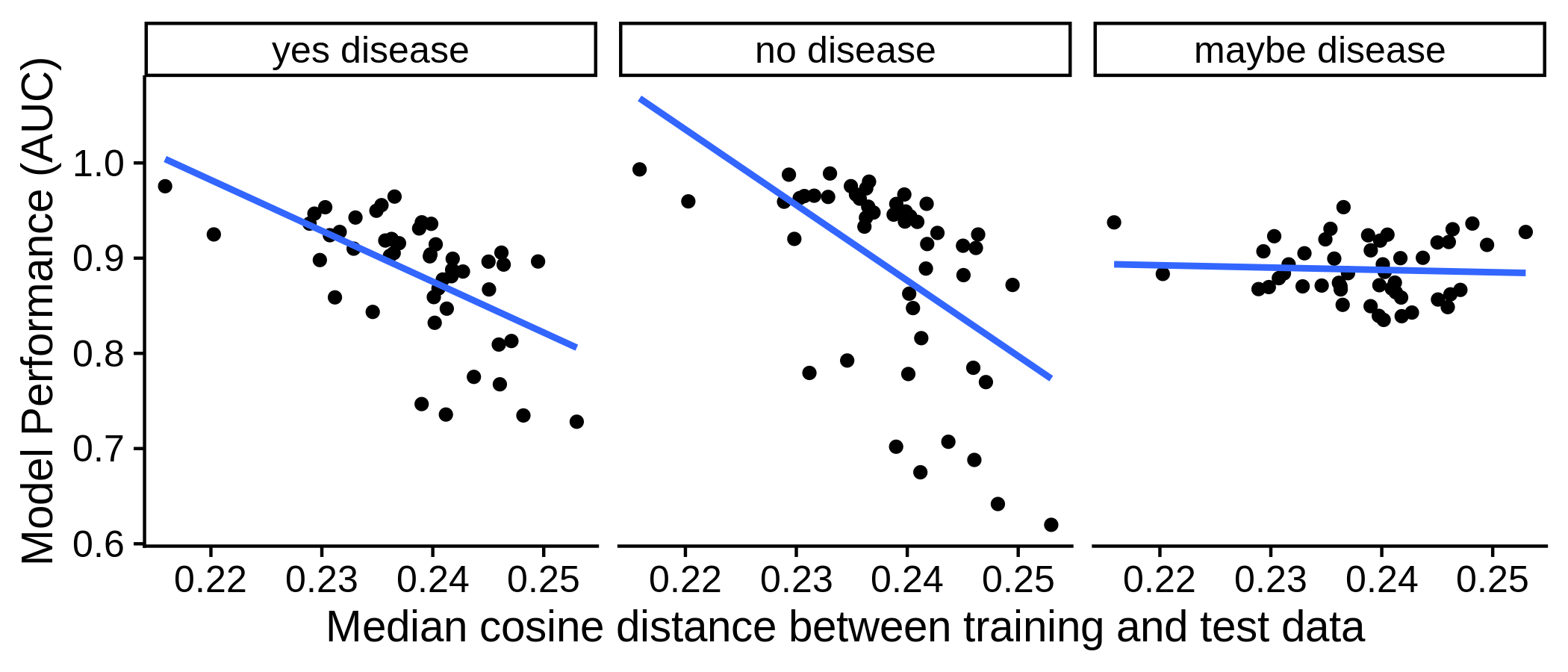}
    \caption{AUC vs. semantic similarity between a model’s training and test datasets.  Semantic similarity is measured with median cosine distance.  Each point represents one performance test (train-test pair) for each label.  Trend lines are fitted using OLS.\\  
}
    \label{fig3}
\end{figure}

\begin{table}[hbt!]
\centering
\caption{Ordinary Least-squares regression statistics}
    \begin{tabular}{|l|r|r|r|}
    \hline
    Diagnosis Label &  Slope &  $R^2$ &   p-value \\ \hline
    
              ``Yes" &   -5.3 &      0.354 &  0.000006 \\ \hline
               ``No" &   -7.9 &      0.301 &  0.000045 \\ \hline
            ``Maybe" &   -0.2 &      0.003 &  0.700000 \\ \hline
    
    \end{tabular}
\end{table}

\subsection*{Alternative Choice of Performance Measure}
We also investigated the use of a precision (also known as positive predictive value or PPV) as a performance measure. For each train-test set pair, we measured PPV at a recall cutoff of 0.9. We repeated the analyses in the Model Performance and Secondary Performance Evaluation sections using PPV rather than AUC (see Supplementary Figures 6-7 and Supplementary Table 7).

\section*{Discussion}
\section*{Comparison with Previous Studies}
Generalizability continues to be one of the most important concerns in biomedical natural language processing, since a resolution of the generalizability problem would allow for widespread deployment of NLP models to both research and clinical practice. Previous studies have highlighted the difficulty of making generalizable NLP models in applications such as, negation detection, cancer diagnosis, and temporal reasoning (Wu et al, Santus et al, Velupillai et al respectively)\cite{ref10, ref7, ref8}.

Wu et. al examined whether a rule-based negation detection model could generalize to clinical text corpora on which it was not trained\cite{ref10}. The authors established that while a model could be optimized to increase F1-score on an individual corpus of text, that model could not then be deployed on another corpus with comparable performance without first undergoing domain adaptation. 

Santus et al investigated the generalizability of a breast cancer diagnosis model across institutions. Their model used a convolutional neural network (CNN) coupled with a logistic regression unit to predict the presence of breast cancer based on a medical note (represented as a matrix of sentence embeddings)\cite{ref7}. Santus et. al. demonstrated that increasing the number of hospitals represented in the training set improved accuracy on test data from unseen institutions.

Our study improved upon the work from Santus et al\cite{ref7} in several ways. We used a state of the art NLP classification architecture (sciBERT), designed training and test cohorts in a combinatorial fashion, and used more refined performance metrics and distance measures between datasets. A BERT-based architecture allowed our model to consider the relationship between any two tokens in an input sentence as opposed to older language models. We used AUC as a performance measure rather than accuracy since AUC can be computed separately for each class and measures model performance across classification thresholds. The prior state-of-the-art NLP generalizability study reports that combining 2 or 3 datasets increases ``diversity" and often improves performance on external tests.  They only used one pair of hospitals to assess the effect of combining datasets.  We trained models on all combinations of 2 (or all 3) medical specialties to better estimate model performance trends and computed group-comparison statistics.  We use MCDs (which are mentioned descriptively in Santus et al\cite{ref1}. to explain and statistically test differences in model generalization across this continuous measure of train-test dataset difference.  Furthermore, we report the deeper insight that generalization trends are further conditioned by the specific label in a classifier’s possible outputs.  This study found that generalization is even an issue within a single critical care unit when models are tested on new medical specialties, and further found factors that are significantly associated with how well a model will generalize, and resolved finer differences between specific labels’ generalizability.

We defined ``internal", ``partial", and ``external" relations between a training set and test sets to investigate generalizability by medical specialty. Combinatorially designed cohorts and train-test relations made it possible to use the repeated measures ANOVA test to determine statistical significance of the differences in average AUC for both generalizability by specialty and train-test set BERT-embedding differences, which previous studies had not considered.
\subsection*{Unsupervised Result Interpretation}
We used tSNE and PCA to examine whether the sciBERT embeddings would cluster according to some property of the corresponding sentences/notes.

tSNE demonstrated that both note and sentence embeddings clustered based on the type of EHR note (Figure 1A, B), but is markedly more pronounced at the document level.  This could be explained by the smaller size of individual sentences compared to entire notes, and is likely reinforced by different medical specialties using different sets of EHR note templates.  

We found that sentences/notes clustered in the tSNE projection by note type rather than medical specialty. This is consistent with the notion that different note types require different vocabulary and may be written by different types of caregivers (e.g respiratory therapists for respiratory notes, nurses for nursing notes). 

PCA confirmed the notion that note type is the primary signal in clustering sentence embeddings (Figure 1D, 1E). In addition, cardiology and pulmonology sentences are more similarly distributed in the PCA projection than either is to oncology sentences. This is not unexpected since cardiac and pulmonary conditions share underlying physiology and often present together in ICU patients.
\subsubsection*{Supervised Result Interpretation}
Using the repeated-measures ANOVA test on the single-specialty models, we were able to show that the differences in macro-averaged AUC between internal, partial overlap, and external test sets were statistically significant as shown in Figure 2A (p = 0.016). Model performance was thus found to decrease when the test set had a smaller percentage of sentences from the same specialty as those in the training set. This is a confirmation of the generalizability issue at the medical specialty level, since models trained on a single specialty could not achieve comparable performance on test sets containing sentences from unseen specialties. We note that the relatively high performance of oncology-trained models in Figure 2A may be due to an overrepresentation of oncology articles in the corpora used to pretrain sciBERT compared to articles on other specialties.

However, we have also found that increasing the number of specialties represented in the training set upfront can improve model generalizability on unseen test specialties. Thus, our result corroborates the central finding from Santus et al\cite{ref7} and is also supported by a statistical significance test. Unlike Santus\cite{ref7}, we created specialties for train and test sets in a combinatorial fashion, providing a more comprehensive view of the effect of training set diversity. Our finding suggests that a general purpose diagnosis model might have to be fine-tuned using a heterogenous training set to allow the model to generalize on a wider range of test sets.

We showed that train set - test set similarity is positively correlated with increased classification performance for sentences with positive and negative sentiment (``Yes" and ``No" ground truth labels). In general, greater train-test distance resulted in lower classification AUC score for ``Yes" and ``No" labels but not for ``Maybe" labels (Figure 3).  Thus, it is easier to generalize on ``Maybe" sentences from a dissimilar test set since there is relatively small loss of AUC compared to ``Yes" and ``No" sentences.

We believe that cosine distance between the unsupervised embeddings of datasets provides invaluable information for machine learning practitioners during model deployment and development. In model deployment, a practitioner can assess whether a model can generalize to a new dataset without requiring any labelled data. Similarly, during iterative model development, the practitioner can measure the distance between a new, unlabeled dataset and an existing training set to ensure maximal gain in data diversity at every iterative step of model development. In instances where it is expensive to acquire labelled data, such as in healthcare settings, this approach will greatly reduce the time and resources required to develop robust NLP models.
\subsection*{Limitations}
Our analysis is limited by our sentence collection methodology, performance measures, and use of MCD as a proxy for dataset similarity. 

The oncology sentences had a disproportionately large percentage of discharge summary sentences - 68\% of oncology sentences were from discharge summaries whereas discharge summaries made up only 3\% of all notes in the MIMIC-III NOTEEVENTS table. Additionally, the oncology-only test set contained only discharge summary sentences since test sentences were tagged solely by the resident physician annotator. We had previously determined that note type was the primary signal in the tSNE projection. Thus, note type is likely the primary signal in the high-dimensional embedding as well, since tSNE preserves local relationships. Having an oncology set enriched for discharge sentences could have allowed models trained on oncology-containing test sets to perform with unusually high AUC on oncology-containing test sets, whereas models fine-tuned on cardiology and pulmonology sentences would have to classify sentences from a greater diversity of note types correctly. Another concern with our sentence extraction approach is that MIMIC-III is an ICU-specific dataset and contains notes only from specialties and indications that are represented in critical care. A dataset containing sentences from all departments of the hospital would provide a larger pool of sentences to sample from for model training and testing. 

Our choice of AUC as a performance metric may not be ideal for use in a real-time clinical setting where identifying patients with severe or rare diseases is a major concern for doctors. We selected AUC due to its multiple advantages in measuring classifier performance over accuracy, though in a clinical setting positive predictive value (PPV, also known as precision) may be more important in order to correctly identify patients who have a disease. In this case, one could define a PPV cutoff based on the particular classification task at hand (as we did in the supplementary results section) or use the average precision derived from precision-recall curves (see Supplementary Figures 3-4). 

We established that MCD provided a reasonable proxy of dataset similarity. We found statistically significant differences in mean MCD between internal, partial, and external train-test pairs. Our finding that the difference in mean MCD between internal train-test pairs and intra-dataset pairs was not unexpected since the original train-test split was random. However, MCD may not recapitulate dataset similarity properly for all combinations of train and test sets. For example, a train set-test set pair where 51\% of sentence pairs have cosine distance 0 and 49\% of sentence pairs have cosine distance of 2 is judged to be identical via MCD, while a train set-test set pair where 49\% of sentence pairs are identical and 51\% of sentence pairs have cosine distance of 2 is judged to be completely different. A single statistic may not adequately capture dataset similarity in all cases; a collection of metrics may be more descriptive of dataset similarity.

A final point to note is that the relationship between AUC and MCD may not be best explained with a linear model for a wider range of values. An analysis using a larger number of medical specialties or a covariate with markedly different embedding distributions (e.g. different note types) may reveal that a different model may show a stronger association between AUC and MCD compared to the linear model.

\subsection*{Future Directions}
Follow-up studies could include assessing generalizability of specialty sentences along other metadata axes from the MIMIC-III database, such as note type or patient demographics. Additionally, one could ask other questions about the sentences collected for this study, such as patient prognosis given a sentence and the name of a disease contained in that sentence. One could also examine the effect of using different pretrained BERT models, such as BioBERT or an EHR-pretrained BERT on AUCs and generalizability.

\section*{Acknowledgements}
We’d like to acknowledge support for this work from Nference.  The healthcare AI company has provided fantastic apps to accelerate this research on making biopharmaceutical and clinical data more computable.

We'd like to thank Joseph Lehar for reviewing the manuscript and providing feedback.

\section*{Authors' Disclosures of Conflicts of Interest}
Marcus Badgeley: Employee of Neurable at the time of writing.

\makeatletter
\renewcommand{\@biblabel}[1]{\hfill #1.}
\makeatother

\bibliographystyle{unsrt}

\section*{Supplementary Results}

\newcommand{\beginsupplement}{%
        \setcounter{table}{0}
        \renewcommand{\thetable}{S\arabic{table}}%
        \setcounter{figure}{0}
        \renewcommand{\thefigure}{S\arabic{figure}}%
     }
\beginsupplement

\subsubsection*{Supplementary Tables}

\FloatBarrier
\begin{table}[!htb]
\centering
\caption{Distribution of MIMIC Documents and Specialty Sentences (Train + Test) by Label and Note Type
}
    \begin{tabular}{|l|r|r|r|r|}
    \hline
                                       Table Name &  Document-Level tSNE notes &  Oncology  &  Cardiology  &  Pulmonology \\ \hline
                              Number of Documents &                     5020.0 &                  752 &                   760 &                     788 \\ \hline
                              Number of Sentences &                   102705.0 &                 1120 &                   902 &                     933 \\ \hline
                            Yes-Labeled Sentences &                        N/A &                  796 &                   630 &                     457 \\ \hline
                             No-Labeled Sentences &                        N/A &                  147 &                   167 &                     161 \\ \hline
                          Maybe-Labeled Sentences &                        N/A &                  177 &                   105 &                     315 \\ \hline
     Note Type - Nursing/other &                     2000.0 &                  141 &                   443 &                     456 \\ \hline
         Note Type - Radiology &                     1250.0 &                  120 &                   259 &                     286 \\ \hline
           Note Type - Nursing &                      500.0 &                   26 &                    15 &                      24 \\ \hline
               Note Type - ECG &                      500.0 &                    0 &                    89 &                      52 \\ \hline
         Note Type - Physician &                      350.0 &                   37 &                    19 &                      26 \\ \hline
 Note Type - Discharge summary &                      150.0 &                  764 &                    30 &                      37 \\ \hline
              Note Type - Echo &                      110.0 &                    9 &                    21 &                      13 \\ \hline
       Note Type - Respiratory &                       75.0 &                    2 &                    10 &                      14 \\ \hline
         Note Type - Nutrition &                       25.0 &                    3 &                     4 &                       5 \\ \hline
           Note Type - General &                       20.0 &                    6 &                     4 &                       4 \\ \hline
    Note Type - Rehab Services &                       15.0 &                    3 &                     2 &                       5 \\ \hline
       Note Type - Social Work &                       10.0 &                    7 &                     2 &                       3 \\ \hline
   Note Type - Case Management &                        5.0 &                    2 &                     0 &                       1 \\ \hline
          Note Type - Pharmacy &                        5.0 &                    0 &                     1 &                       2 \\ \hline
           Note Type - Consult &                        5.0 &                    0 &                     3 &                       5 \\ \hline
\end{tabular}
\end{table}

\begin{table}[!htb]
\centering
\caption{Median cosine distance and AUC for each train-test pair by class
}
\begin{tabular}{|l|l|r|r|r|r|} \hline
TRAIN SET NAME &           TEST SET NAME &  MEDIAN COSINE DISTANCE &   AUC\_YES &    AUC\_NO &  AUC\_MAYBE \\ \hline
cancer train &             cancer test &                0.215871 &  0.975534 &  0.993217 &   0.937577 \\ \hline
cancer train &            cardiac test &                0.249490 &  0.896512 &  0.871857 &   0.913924 \\ \hline
cancer train &          pulmonary test &                0.241795 &  0.899403 &  0.914844 &   0.839045 \\ \hline
cancer train &     cancer cardiac test &                0.230304 &  0.953472 &  0.963071 &   0.923028 \\ \hline
cancer train &   cancer pulmonary test &                0.228892 &  0.936262 &  0.959265 &   0.867502 \\ \hline
cancer train &  cardiac pulmonary test &                0.246187 &  0.905773 &  0.910722 &   0.861978 \\ \hline
cancer train &          all three test &                0.236277 &  0.920109 &  0.942543 &   0.870597 \\ \hline
cardiac train &             cancer test &                0.252976 &  0.728205 &  0.619940 &   0.927469 \\ \hline
cardiac train &            cardiac test &                0.240513 &  0.868217 &  0.847582 &   0.924684 \\ \hline
cardiac train &          pulmonary test &                0.241746 &  0.887863 &  0.957079 &   0.858635 \\ \hline
cardiac train &     cancer cardiac test &                0.246044 &  0.767515 &  0.688085 &   0.916972 \\ \hline
cardiac train &   cancer pulmonary test &                0.247087 &  0.812982 &  0.769822 &   0.866655 \\ \hline
cardiac train &  cardiac pulmonary test &                0.240901 &  0.877588 &  0.938111 &   0.868122 \\ \hline
cardiac train &          all three test &                0.245950 &  0.809263 &  0.784871 &   0.848585 \\ \hline
pulmonary train &             cancer test &                0.241182 &  0.735755 &  0.675118 &   0.874306 \\ \hline
pulmonary train &            cardiac test &                0.240172 &  0.832115 &  0.862669 &   0.835127 \\ \hline
pulmonary train &          pulmonary test &                0.220262 &  0.924925 &  0.959641 &   0.883404 \\ \hline
pulmonary train &     cancer cardiac test &                0.238994 &  0.746759 &  0.701956 &   0.849562 \\ \hline
pulmonary train &   cancer pulmonary test &                0.231179 &  0.858818 &  0.779460 &   0.883988 \\ \hline
pulmonary train &  cardiac pulmonary test &                0.229819 &  0.897995 &  0.920222 &   0.869629 \\ \hline
pulmonary train &          all three test &                0.234585 &  0.843544 &  0.792489 &   0.871215 \\ \hline
cancer cardiac train &             cancer test &                0.233030 &  0.942668 &  0.988911 &   0.905116 \\ \hline
cancer cardiac train &            cardiac test &                0.245067 &  0.867143 &  0.882177 &   0.856561 \\ \hline
cancer cardiac train &          pulmonary test &                0.240264 &  0.914515 &  0.943946 &   0.885536 \\ \hline
cancer cardiac train &     cancer cardiac test &                0.236957 &  0.915702 &  0.947896 &   0.884223 \\ \hline
cancer cardiac train &   cancer pulmonary test &                0.236298 &  0.919898 &  0.973313 &   0.867163 \\ \hline
cancer cardiac train &  cardiac pulmonary test &                0.242719 &  0.885958 &  0.926500 &   0.842869 \\ \hline
cancer cardiac train &          all three test &                0.239733 &  0.901924 &  0.966845 &   0.839409 \\ \hline
cancer pulmonary train &             cancer test &                0.229330 &  0.946759 &  0.987672 &   0.907176 \\ \hline
cancer pulmonary train &            cardiac test &                0.246388 &  0.893245 &  0.924855 &   0.930380 \\ \hline
cancer pulmonary train &          pulmonary test &                0.232867 &  0.909965 &  0.964309 &   0.870354 \\ \hline
cancer pulmonary train &     cancer cardiac test &                0.235716 &  0.918519 &  0.962478 &   0.899522 \\ \hline
cancer pulmonary train &   cancer pulmonary test &                0.230722 &  0.924093 &  0.965221 &   0.878978 \\ \hline
cancer pulmonary train &  cardiac pulmonary test &                0.239785 &  0.903910 &  0.938444 &   0.871589 \\ \hline
cancer pulmonary train &          all three test &                0.236476 &  0.905279 &  0.954077 &   0.850971 \\ \hline
cardiac pulmonary train &             cancer test &                0.248168 &  0.734793 &  0.641796 &   0.936343 \\ \hline
cardiac pulmonary train &            cardiac test &                0.241680 &  0.881008 &  0.889072 &   0.900000 \\ \hline
cardiac pulmonary train &          pulmonary test &                0.231606 &  0.927532 &  0.965590 &   0.893466 \\ \hline
cardiac pulmonary train &     cancer cardiac test &                0.243704 &  0.775309 &  0.707172 &   0.900398 \\ \hline
cardiac pulmonary train &   cancer pulmonary test &                0.240093 &  0.859105 &  0.778323 &   0.893500 \\ \hline
cardiac pulmonary train &  cardiac pulmonary test &                0.236136 &  0.902431 &  0.933111 &   0.874114 \\ \hline
cardiac pulmonary train &          all three test &                0.241261 &  0.846957 &  0.815987 &   0.864179 \\ \hline
all three train &             cancer test &                0.236553 &  0.964672 &  0.980351 &   0.953549 \\ \hline
all three train &            cardiac test &                0.245023 &  0.896346 &  0.913056 &   0.916456 \\ \hline
all three train &          pulmonary test &                0.234916 &  0.949796 &  0.975611 &   0.919744 \\ \hline
all three train &     cancer cardiac test &                0.238771 &  0.931211 &  0.945702 &   0.923984 \\ \hline
all three train &   cancer pulmonary test &                0.235375 &  0.955649 &  0.966858 &   0.930806 \\ \hline
all three train &  cardiac pulmonary test &                0.239856 &  0.936058 &  0.949000 &   0.918438 \\ \hline
all three train &          all three test &                0.239009 &  0.937647 &  0.956921 &   0.908204 \\ \hline
\end{tabular}
\end{table}

\begin{table}[!htb]
\centering
\caption{Intra-Dataset median cosine distances}
    \begin{tabular}{|l|r|}\hline
                 DATASET &  INTRA-DATASET MEDIAN COSINE DISTANCE \\ \hline
            cancer train &                              0.222537 \\ \hline
           cardiac train &                              0.240437 \\ \hline
         pulmonary train &                              0.221264 \\ \hline
    cancer cardiac train &                              0.240346 \\ \hline
  cancer pulmonary train &                              0.233522 \\ \hline
 cardiac pulmonary train &                              0.238165 \\ \hline
         all three train &                              0.240545 \\ \hline
             cancer test &                              0.202743 \\ \hline
            cardiac test &                              0.237297 \\ \hline
          pulmonary test &                              0.218192 \\ \hline
     cancer cardiac test &                              0.229778 \\ \hline
   cancer pulmonary test &                              0.227121 \\ \hline
  cardiac pulmonary test &                              0.234084 \\ \hline
          all three test &                              0.237062 \\ \hline
\end{tabular}
\end{table}

\begin{table}[!htb]
\centering
\caption{ Ten Most Common Indications by Specialty (Number Sentences)}
    \begin{tabular}{|r|l|l|l|} \hline

Rank &              Oncology &                  Cardiology &                                 Pulmonology \\ \hline

          1 &            cyst (118) &           tachycardia (308) &                             pneumonia (250) \\ \hline
          2 &    breast cancer (97) &              aneurysm (149) &                          pneumothorax (145) \\ \hline
          3 &         lymphoma (82) &               ischemia (81) &                           atelectasis (126) \\ \hline
          4 &      lung cancer (73) &          heart failure (58) &                      pleural effusion (107) \\ \hline
          5 &            cysts (56) &           endocarditis (47) &                        pulmonary edema (68) \\ \hline
          6 &       hemangioma (44) &           cardiomegaly (41) &                    respiratory failure (51) \\ \hline
          7 &         melanoma (39) &         cardiomyopathy (27) &                                 asthma (47) \\ \hline
          8 &  prostate cancer (38) &    atrial fibrillation (27) &                                   copd (38) \\ \hline
          9 &     colon cancer (35) &         cardiac arrest (27) &  chronic obstructive pulmonary disease (36) \\ \hline
         10 &   adenocarcinoma (34) &  myocardial infarction (26) &                              emphysema (28) \\ \hline
\end{tabular}
\end{table}

\begin{table}[!htb]
\caption{Median cosine distance and PPV (at recall = 0.9) for each train-test pair by class}
    \begin{tabular}{|l|l|r|r|r|r|} \hline

          TRAIN SET NAME &           TEST SET NAME &  MEDIAN COSINE DISTANCE &   PPV\_YES &    PPV\_NO &  PPV\_MAYBE \\ \hline

            cancer train &             cancer test &                0.215871 &  0.967742 &  0.941176 &   0.253012 \\ \hline
            cancer train &            cardiac test &                0.249490 &  0.941176 &  0.218391 &   0.227848 \\ \hline
            cancer train &          pulmonary test &                0.241795 &  0.770270 &  0.382609 &   0.459459 \\ \hline
            cancer train &     cancer cardiac test &                0.230304 &  0.950980 &  0.525424 &   0.255814 \\ \hline
            cancer train &   cancer pulmonary test &                0.228892 &  0.880240 &  0.591549 &   0.436620 \\ \hline
            cancer train &  cardiac pulmonary test &                0.246187 &  0.870370 &  0.349515 &   0.422535 \\ \hline
            cancer train &          all three test &                0.236277 &  0.883333 &  0.486111 &   0.390625 \\ \hline
           cardiac train &             cancer test &                0.252976 &  0.853061 &  0.131222 &   0.328125 \\ \hline
           cardiac train &            cardiac test &                0.240513 &  0.946078 &  0.089623 &   0.200000 \\ \hline
           cardiac train &          pulmonary test &                0.241746 &  0.745098 &  0.637681 &   0.565517 \\ \hline
           cardiac train &     cancer cardiac test &                0.246044 &  0.858407 &  0.132479 &   0.323529 \\ \hline
           cardiac train &   cancer pulmonary test &                0.247087 &  0.765625 &  0.181034 &   0.370588 \\ \hline
           cardiac train &  cardiac pulmonary test &                0.240901 &  0.825581 &  0.461538 &   0.464000 \\ \hline
           cardiac train &          all three test &                0.245950 &  0.795000 &  0.151261 &   0.304348 \\ \hline
         pulmonary train &             cancer test &                0.241182 &  0.867769 &  0.162304 &   0.148936 \\ \hline
         pulmonary train &            cardiac test &                0.240172 &  0.906103 &  0.208791 &   0.093750 \\ \hline
         pulmonary train &          pulmonary test &                0.220262 &  0.820144 &  0.637681 &   0.551282 \\ \hline
         pulmonary train &     cancer cardiac test &                0.238994 &  0.850877 &  0.160622 &   0.165414 \\ \hline
         pulmonary train &   cancer pulmonary test &                0.231179 &  0.812155 &  0.218085 &   0.398734 \\ \hline
         pulmonary train &  cardiac pulmonary test &                0.229819 &  0.825581 &  0.327103 &   0.375000 \\ \hline
         pulmonary train &          all three test &                0.234585 &  0.823834 &  0.189189 &   0.318471 \\ \hline
    cancer cardiac train &             cancer test &                0.233030 &  0.949541 &  0.727273 &   0.202020 \\ \hline
    cancer cardiac train &            cardiac test &                0.245067 &  0.940299 &  0.169643 &   0.122137 \\ \hline
    cancer cardiac train &          pulmonary test &                0.240264 &  0.808333 &  0.457447 &   0.549020 \\ \hline
    cancer cardiac train &     cancer cardiac test &                0.236957 &  0.937198 &  0.534483 &   0.174603 \\ \hline
    cancer cardiac train &   cancer pulmonary test &                0.236298 &  0.879518 &  0.646154 &   0.351955 \\ \hline
    cancer cardiac train &  cardiac pulmonary test &                0.242719 &  0.871166 &  0.387097 &   0.352941 \\ \hline
    cancer cardiac train &          all three test &                0.239733 &  0.893258 &  0.493151 &   0.268817 \\ \hline
  cancer pulmonary train &             cancer test &                0.229330 &  0.967742 &  0.842105 &   0.218750 \\ \hline
  cancer pulmonary train &            cardiac test &                0.246388 &  0.941463 &  0.306452 &   0.305085 \\ \hline
  cancer pulmonary train &          pulmonary test &                0.232867 &  0.820144 &  0.758621 &   0.524390 \\ \hline
  cancer pulmonary train &     cancer cardiac test &                0.235716 &  0.937198 &  0.632653 &   0.271605 \\ \hline
  cancer pulmonary train &   cancer pulmonary test &                0.230722 &  0.885542 &  0.724138 &   0.346154 \\ \hline
  cancer pulmonary train &  cardiac pulmonary test &                0.239785 &  0.855422 &  0.486486 &   0.344828 \\ \hline
  cancer pulmonary train &          all three test &                0.236476 &  0.878453 &  0.692308 &   0.255102 \\ \hline
 cardiac pulmonary train &             cancer test &                0.248168 &  0.878661 &  0.159763 &   0.218750 \\ \hline
 cardiac pulmonary train &            cardiac test &                0.241680 &  0.941463 &  0.158333 &   0.136364 \\ \hline
 cardiac pulmonary train &          pulmonary test &                0.231606 &  0.883721 &  0.785714 &   0.597222 \\ \hline
 cardiac pulmonary train &     cancer cardiac test &                0.243704 &  0.885845 &  0.161677 &   0.170543 \\ \hline
 cardiac pulmonary train &   cancer pulmonary test &                0.240093 &  0.847953 &  0.207254 &   0.386503 \\ \hline
 cardiac pulmonary train &  cardiac pulmonary test &                0.236136 &  0.898734 &  0.360000 &   0.368750 \\ \hline
 cardiac pulmonary train &          all three test &                0.241261 &  0.850267 &  0.189189 &   0.284091 \\ \hline
         all three train &             cancer test &                0.236553 &  0.967593 &  0.533333 &   0.403846 \\ \hline
         all three train &            cardiac test &                0.245023 &  0.955446 &  0.387755 &   0.206897 \\ \hline
         all three train &          pulmonary test &                0.234916 &  0.876923 &  0.800000 &   0.618705 \\ \hline
         all three train &     cancer cardiac test &                0.238771 &  0.941748 &  0.484375 &   0.244444 \\ \hline
         all three train &   cancer pulmonary test &                0.235375 &  0.907407 &  0.512195 &   0.512195 \\ \hline
         all three train &  cardiac pulmonary test &                0.239856 &  0.898734 &  0.571429 &   0.475806 \\ \hline
         all three train &          all three test &                0.239009 &  0.898305 &  0.507042 &   0.426087 \\ \hline

\end{tabular}
\end{table}

\begin{table}[!htb]
\centering
\caption{ OLS results for PPV vs Median Cosine Distance}
    \begin{tabular}{|l|r|r|r|} \hline

Diagnosis Label &      Slope &  R\textasciicircum 2 value &       p-value \\ \hline

          ``Yes" &  -1.165593 &      0.021 &  3.170000e-01 \\ \hline
           ``No" & -22.372893 &      0.472 &  5.100000e-08 \\ \hline
        ``Maybe" &  -4.455773 &      0.056 &  1.020000e-01 \\ \hline

\end{tabular}
\end{table}
\FloatBarrier
\FloatBarrier
\subsubsection*{Supplementary Figures - Model Performance}

\begin{figure}[hbt!]
    \centering
    \includegraphics[width=\linewidth]{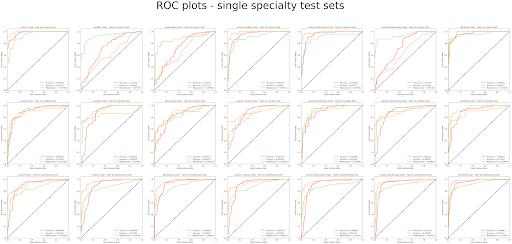}
    \caption{ROC plots for models trained on 7 datasets with mixed specialty compositions and tested each of 3 specialties.  Each row represents a different test set and each column represents a different model (training set).
}
    \label{fig4}
\end{figure}

\begin{figure}[hbt!]
    \centering
    \includegraphics[width=\linewidth]{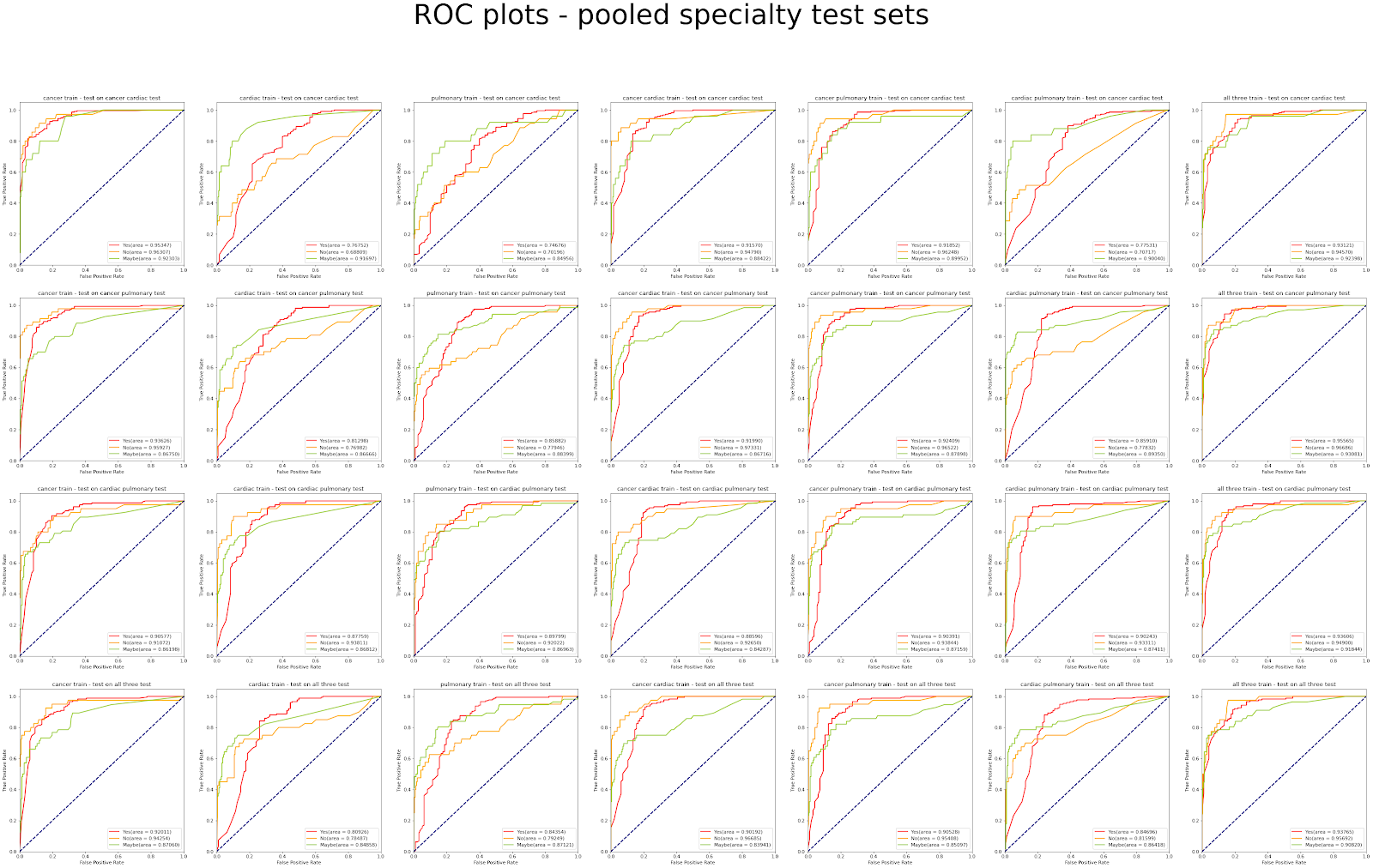}
    \caption{ROC plots for pooled disease class test sets, evaluated on all seven models.  As in Figure S1, each row represents a different test set and each column represents a different model.
}
    \label{fig5}
\end{figure}

\begin{figure}[hbt!]
    \centering
    \includegraphics[width=\linewidth]{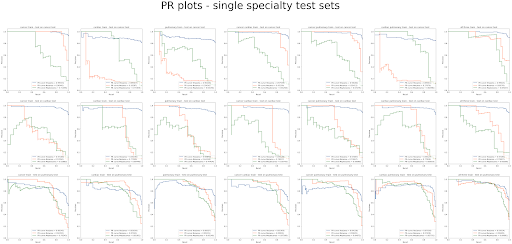}
    \caption{PR curves for single disease class test sets, evaluated on all seven models}
    \label{fig6}
\end{figure}

\begin{figure}[hbt!]
    \centering
    \includegraphics[width=\linewidth]{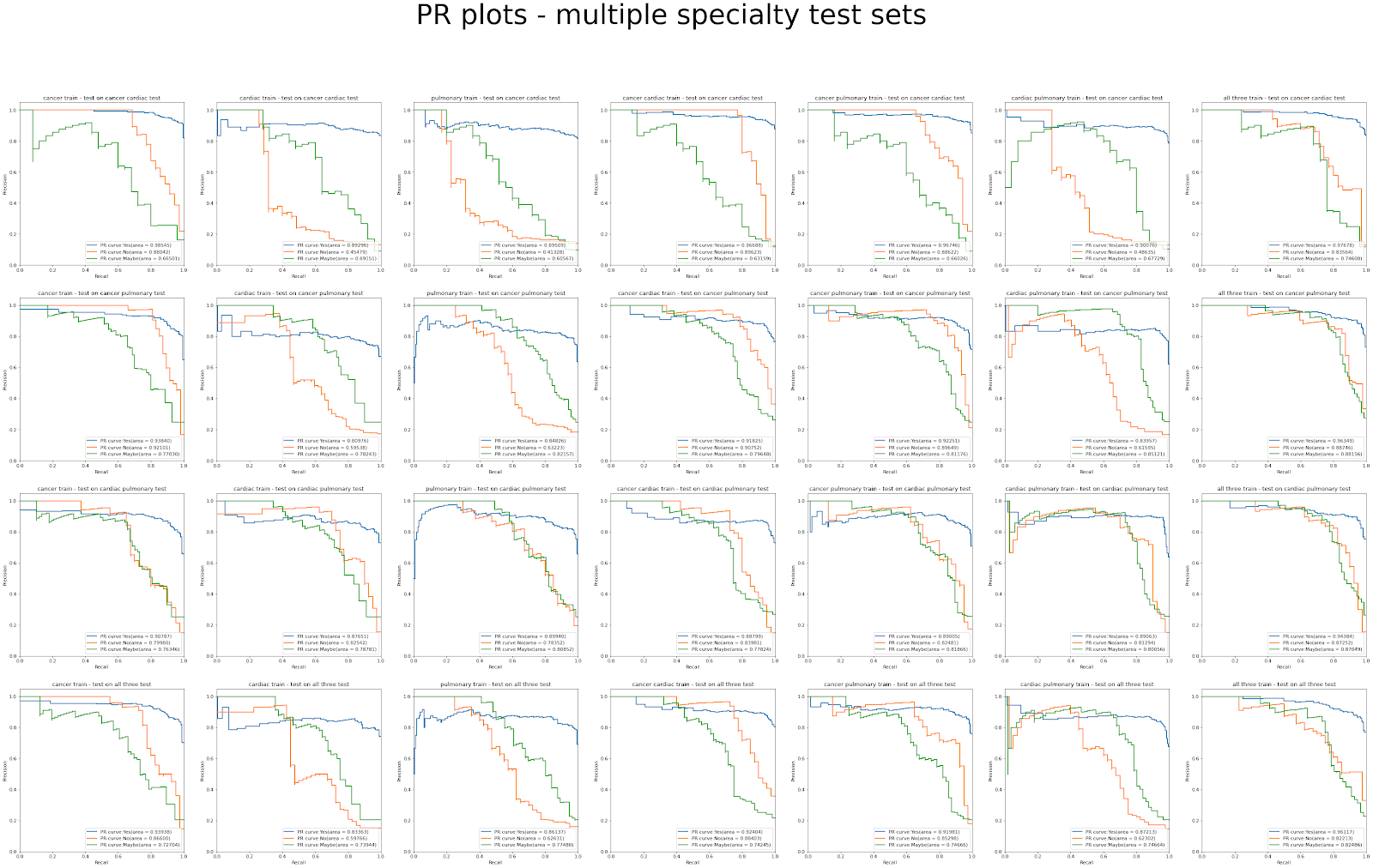}
    \caption{PR curves for the pooled disease class test sets, evaluating all seven models.}
    \label{fig7}
\end{figure}
\FloatBarrier

\subsubsection*{Supplementary Figures - Dataset Characteristics}
\FloatBarrier
\begin{figure}
    \begin{subfigure}{.75\linewidth}
    \centering
    \includegraphics[width=.75\linewidth]{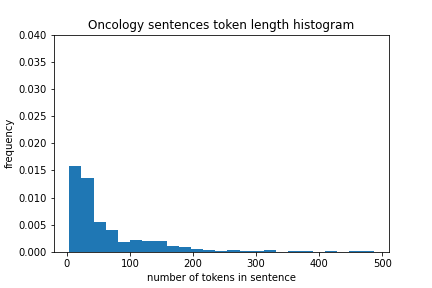}
    \end{subfigure}
    \begin{subfigure}{.75\linewidth}
    \centering
    \includegraphics[width=.75\linewidth]{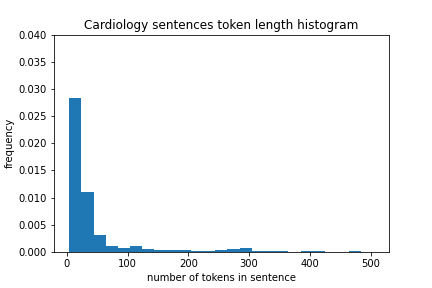}
    \end{subfigure}
    \newline
    \newline
    \newline
    \begin{subfigure}{.75\linewidth}
    \centering
    \includegraphics[width=.75\linewidth]{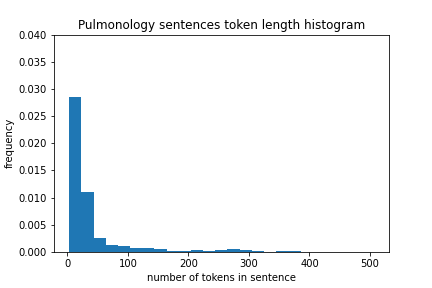}
    \end{subfigure}
    
    \label{fig8}
    \caption{Histogram of sentence token length for each medical specialty. Oncology sentences are on average longer than cardiology and pulmonology sentences. While cardiology and pulmonology sentences have comparable median token length (22 and 20 tokens respectively), cardiology sentences have a larger standard deviation of token length (75.5 tokens vs 66.5 tokens).}
\end{figure}
\FloatBarrier
\subsubsection*{Supplementary Figures - Alternate Choice of Performance
Metrics}\FloatBarrier
We investigated whether using PPV (at recall of 0.9) would produce similar results as Figures 2-3 in the main results section, which used AUC.

\begin{figure}[hbt!]
    \centering
    \includegraphics[width=\textwidth]{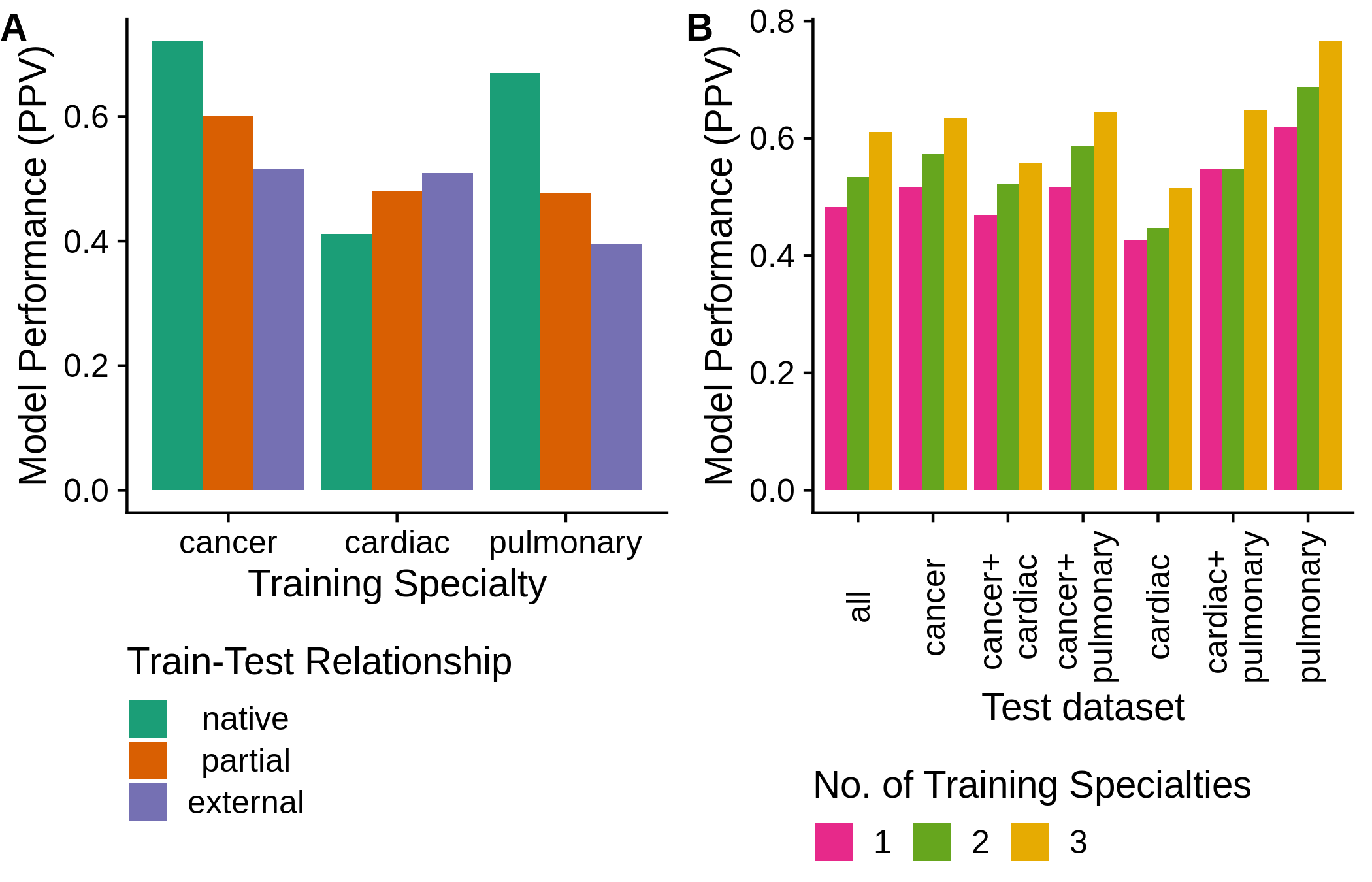}
    \caption{PPV bar plots of model performance tests with different groupings to test primary hypotheses.  (A) PPV vs train-test set relation for each single-disease model, averaged over labels and test sets.  (B) PPV vs how many specialties were represented in the training set.}
    \label{fig9}
\end{figure}

\begin{figure}[hbt!]
    \centering
    \includegraphics[width=\textwidth]{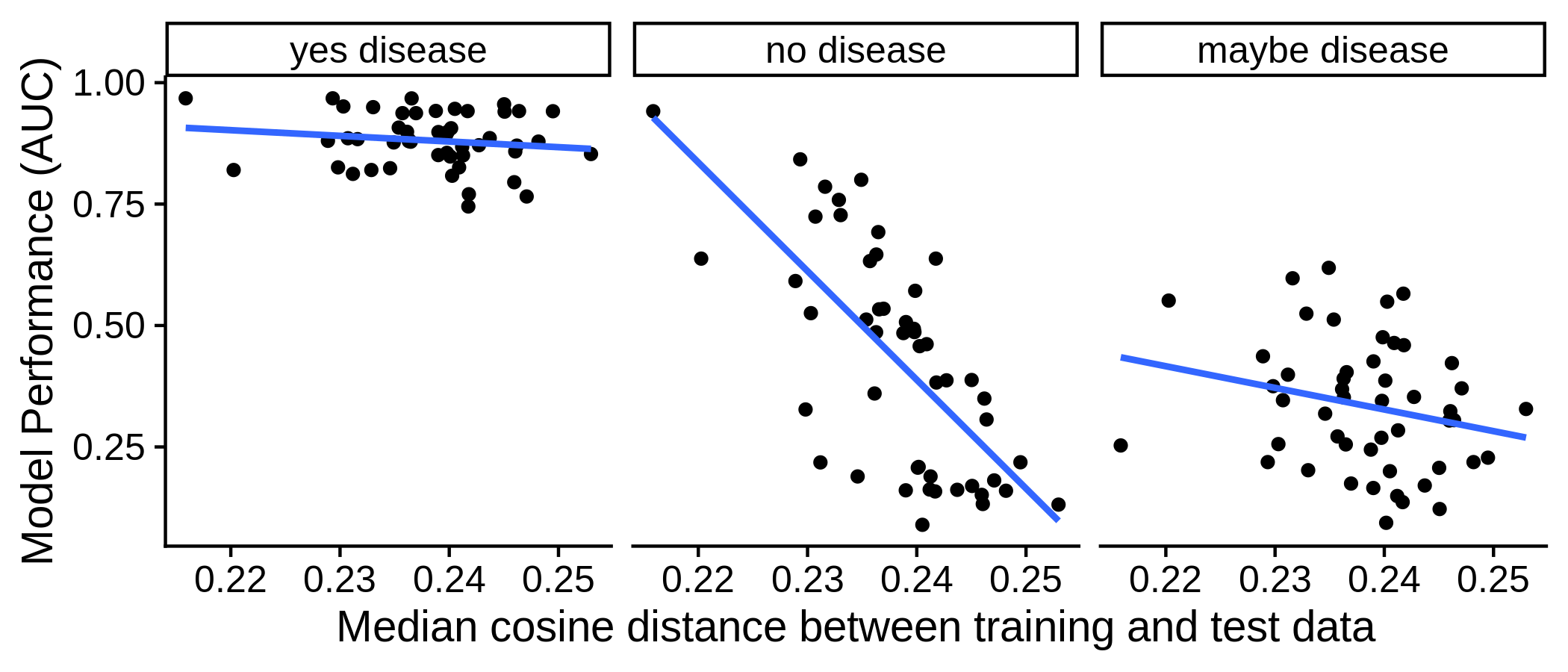}
    \caption{PPV vs. semantic similarity between a model’s training and test datasets.  Semantic similarity is measured with median cosine distance.  Each point represents one performance test (train-test pair) for each label.  Trend lines are fitted using OLS.
}
    \label{fig10}
\end{figure}

\FloatBarrier

\end{document}